\documentclass[10pt, a4paper]{article}
\usepackage{float}  
\usepackage{caption}
\usepackage[utf8]{inputenc}
\usepackage{amsmath}
\usepackage{fontspec}
\usepackage[dvipsnames]{xcolor}
\usepackage{tipa}
\usepackage{fancyhdr}
\usepackage{hyperref}
\usepackage{stfloats}
\usepackage{tabularx}

\newfontfamily\nep{Lohit-Nepali.ttf}

\usepackage[final]{lrec2026} 
\hypersetup{
    colorlinks=true,
    linkcolor=blue,
    filecolor=blue,      
    urlcolor=blue,
    citecolor=blue,           
}

\title{Nwāchā Munā: A Devanagari Speech Corpus and Proximal Transfer Benchmark for Nepal Bhasha ASR}

\name{Rishikesh Kumar Sharma, Safal Narshing Shrestha, Jenny Poudel,\\ {\bf \large Rupak Tiwari, Arju Shrestha, Rupak Raj Ghimire, Bal Krishna Bal}} 

\address{Information and Language Processing Research Lab, Kathmandu University \\
         \{rishi70612, safalnarsingh, jennypoudel100, rupaktiwari18, arzzustha\}@gmail.com,\\
         rughimire@gmail.com, bal@ku.edu.np\\
}

\abstract{
Nepal Bhasha (Newari), an endangered language of the Kathmandu Valley, remains digitally marginalized due to the severe scarcity of annotated speech resources. In this work, we introduce \textbf{Nwāchā Munā} (\textipa{/nwa:tSa: muna:/}), a newly curated 5.39-hour manually transcribed Devanagari speech corpus for Nepal Bhasha, and establish the first benchmark using script-preserving acoustic modeling. We investigate whether proximal cross-lingual transfer from a geographically and linguistically adjacent language (Nepali) can rival large-scale multilingual pretraining in an ultra-low-resource Automatic Speech Recognition (ASR) setting. Fine-tuning a Nepali Conformer model reduces the Character Error Rate (CER) from a 52.54\% zero-shot baseline to 17.59\% with data augmentation, effectively matching the performance of the multilingual \textit{Whisper-Small} model despite utilizing significantly fewer parameters. Our findings demonstrate that proximal transfer from Nepali language serves as a computationally efficient alternative to massive multilingual models. We openly release the dataset and benchmarks to digitally enable the Newari community and foster further research in Nepal Bhasha.
\newline \newline \Keywords{ Automatic Speech Recognition, Nepal Bhasha, Speech Corpus, Low-Resource Speech Recognition}
}

\begin{document}
\pagestyle{empty}

\maketitleabstract

\section{Introduction}
In the current space of AI, high-resource languages dominate in terms of resource availability while indigenous languages remain digitally marginalized. In a country like Nepal, with 124 official languages, this divide is especially vivid, creating a need for technologies like ASR to bridge the gap. Nepal Bhasha (Newari) serves as a prime example of this imbalance, despite being actively spoken by over 860,000 people in Nepal. Genetically, Nepal Bhasha belongs to the Tibeto-Burman branch of the Sino-Tibetan family \cite{ethnologue2026}. Even with a six-century-old history and a designation as one of the official working languages of Bagmati Province \cite{kathmandupost2024bagmati}, Nepal Bhasha is currently classified by UNESCO as a ``definitely endangered'' language \cite{unesco2010atlas}. Building a system such as ASR empowers these communities to maintain their linguistic heritage while fully participating in the modern technological landscape.

The modern field of speech recognition has transitioned from fragmented components based on Hidden Markov Models \cite{rabiner1989tutorial} to unified end-to-end (E2E) architectures such as Transformer \cite{vaswani2017attention} and Conformer \cite{gulati2020conformer}. These modern architectures require large datasets to achieve better performance. Recent advancements have yielded robust ASR systems for languages like Nepali and Hindi, but indigenous languages such as Nepal Bhasha remain severely bottlenecked by an extreme scarcity of annotated speech data. This research addresses the fundamental gap by developing a speech corpus for the Newari language---
Nwāchā Munā. To further demonstrate the usefulness of this dataset, we establish an initial model benchmark and synthesize modern ASR architectures with cross-lingual transfer learning, leveraging a proximate source language to overcome data constraints.

From this research, we aim to provide a scalable blueprint for other underrepresented languages in the region.

Our contributions are threefold:
\begin{enumerate}
    \item We release Nwāchā Munā, a carefully curated 5.39-hour Devanagari speech corpus for Nepal Bhasha.
    \item We provide the first controlled comparison between proximal transfer (Nepali $\rightarrow$ Newari) and multilingual pretraining (Whisper) in this ultra-low-resource setting.
    \item We show that script-preserving proximal transfer can match large multilingual models while requiring substantially fewer parameters and computational resources.
\end{enumerate}

The remainder of this paper is structured as follows: Section 2 reviews related works in low-resource ASR. Section 3 describes the dataset and methodology including data collection and model training strategies. Section 4 presents the experimental results, discussions and error analysis. Section 5 presents the conclusion followed by ethical statement and limitations.

\section{Related works}

One major obstacle in building ASR systems for underrepresented languages is the lack of sufficient speech data. Older methods like Hidden Markov Models required less data but struggled to capture intricate acoustic patterns. With time, new architectures like Transformers \cite{vaswani2017attention} and Conformer \cite{gulati2020conformer} have outperformed older statistical methods, but they require larger datasets. To address this issue, researchers have increasingly turned to transfer learning.
For instance, \citet{cheng2024exploring} demonstrated clear ASR performance gains for two at-risk Austronesian languages, Amis and Seediq, despite having minimal available speech samples. They achieved this by utilizing a novel data-selection scheme to automatically extract phonetically similar utterances from a broad multilingual corpus to augment their training data.
Likewise, significant progress has been made specifically within the South Asian region. The AI4Bharat initiative \cite{javed2022towards} advanced regional ASR by curating over 17,000 hours of raw speech data across 40 Indian languages for continued pre-training.

In the context of Nepali ASR, early efforts relied on Hidden Markov Models (HMM) \cite{ssarma2017hmm} for isolated word recognition, which struggled with continuous speech. The transition to deep learning began with \citet{regmi2019nepali}, they implemented Recurrent Neural Networks (RNN) with Connectionist Temporal Classification (CTC) loss. This was further refined by \citet{bhatta2020nepali} with a hybrid CNN-GRU-CTC model. Furthermore, \citet{regmi-bal-2021-end} implemented end-to-end speech recognition approach for the Nepali language. To address training instability in deeper networks, \citet{dhakal2022automatic} proposed a CNN-BiLSTM-ResNet architecture, achieving a CER of 17.06\%. Concurrently, research began shifting towards data efficiency and cross-lingual techniques. For example, \citet{bansal2020acoustic} explored an acoustic-phonetic approach, demonstrating that leveraging monolingual and cross-lingual information could improve performance in low-resource settings. 
\citet{joshi2023nepali} applied Self-Attention Networks (SAN) to capture long-range dependencies better than recurrent models. Building on these modern architectures, \citet{paudel-etal-2023-large} successfully applied a combination of CNNs and Transformers for large vocabulary continuous speech recognition.

Current state-of-the-art systems have fully embraced E2E Transformer-based architectures. \citet{poudel2025nepconformer} introduced \texttt{NepConformer}, a Conformer-based Nepali ASR model achieving a CER of \texttt{6.01\%}. Recent focus has also turned to data selection, tokenization and resource efficient strategies. \citet{ghimire2023active} proposed an Active Learning framework to select the most informative samples for speech annotation. Another work \cite{ghimire2023pronunciation}, demonstrated that using linguistically motivated sub-word units (syllables) alongside active sampling significantly outperforms standard methods. Work done by \citet{ghimire2025improving} showed that using adapter-based Parameter-Efficient-Fine-Tuning (PEFT) alone improves Word Error Rate (WER) by 19\% compared to \texttt{Whisper-large-v2}. Extending the utility of efficient adaptation, \citet{pantha-etal-2025-speech} demonstrated the effectiveness of PEFT for speech personalization among Nepali speakers. Most recently, \citet{ghimire2023pronunciation} implemented new complementary loss function for addressing  grammatical consistency during model training and fine-tuning.  Their Rule-Based Character Constituency Loss (RBCCL) help to reduce Word Error Rates (WER) from 47.1\% to 23.41\% by penalizing grammatically impossible character sequences during training.

While Nepali has seen consistent growth in the field of ASR, research on Nepal Bhasha remains sparse. A notable exception is the work by \citet{meelen2024end}, who presented the first Newari ASR, where they achieved a CER of 12\%. Their evaluation is conducted in Romanized transliteration, this fails to preserve Devanagari orthography. Their study highlighted that while transfer learning is effective, the lack of standardized orthography and limited data results in higher WER compared to Nepali. They emphasized the need for "orthography standardization" and data augmentation techniques like SpecAugment \cite{park2019specaugment} to make ASR viable for these endangered languages.

Taken together, these findings highlight a persistent gap: prior work lacks ASR systems trained on the Devanagari script using carefully curated low-resource datasets. This research clears these gaps by developing Nwāchā Munā. We utilise Devanagari script for transcribing speech of Nepal Bhasha. We also use data augmentation techniques to train robust ASR models using various cross-lingual transfer strategies and report CER for each technique.

\section{Methodology}
\subsection{Data Collection Technique}
The development of the Nepal Bhasha ASR required a carefully designed data collection process to ensure authenticity and reliability. Both textual and acoustic resources were gathered considering language and speaker diversity. The objective was to construct a dataset that reflects natural language usage while maintaining consistency and structural integrity. The following subsections describe the procedures adopted for textual and audio data acquisition.

\subsubsection{Textual Data Acquisition} 

To establish a baseline for the Nepal Bhasha ASR system, a comprehensive textual corpus was compiled from different sources of digital and print media. The data collection strategy included the use of both formal and everyday language to capture the full range of vocabulary. Digital data were extracted from Nepal Bhasha Wikipedia
\hyperlink{link1}{\textsuperscript{1}} and the OSCAR (Open Super-large Crawled Aggregated coRpus) \cite{ortiz-suarez-etal-2020-monolingual} dataset while print-based resources included regional newspapers, literary manuscripts from the Bloom Library \hyperlink{link2}{\textsuperscript{2}}, and primary school textbooks. In addition to the formal texts obtained, a targeted set of daily conversational sentences was curated manually to incorporate texts that were previously missing. The raw text was rigorously pre-processed where non-target language tokens, numerical digits, and idiosyncratic symbols were manually removed to ensure high fidelity and relevance.  Table \ref{tab:normalization} shows an example of text normalization, where raw text containing Devanagari numerals and symbols is converted into a phonetic, standardized form for acoustic modeling.

\begin{table}[H]
    \centering
    \small
    \caption{Example of Text Normalization for Nepal Bhasha ASR}
    \label{tab:normalization}
    \renewcommand{\arraystretch}{1.6}
    \begin{tabularx}{\columnwidth}{l X} 
        \hline
        \textbf{Condition} & \textbf{Sentence Structure} \\
        \hline
        Before & {\nep''छगु ब्वसय् २१\% ब्व दु।''} \newline 
        chagu bwasay 21\% bwa du \newline
                 \textit{The sentence contains 21 as a number. } \\
        \hline
        After & {\nep छगु ब्वसय् नीछगू सयेक ब्व दु}  \newline 
        chagu bwasay nīchagū sayeka bwa du \newline
                \textit{The sentence contains 21 in textual form } \\
        \hline
    \end{tabularx}
\end{table}

The overall text corpus statistics and sentence length distribution are presented in Table \ref{tab:text_corpus_overview} and \ref{tab:text_corpus_length}:

\thispagestyle{fancy}
\fancyhf{}

\renewcommand{\footrulewidth}{0.4pt} 
\renewcommand{\footrule}{%
    \begin{flushleft}
        \textcolor{gray}{\rule{0.4\columnwidth}{\footrulewidth}} 
    \end{flushleft}
    \vspace{2pt}
}

\renewcommand{\headrulewidth}{0pt} 

\fancyfoot[L]{%
        \textit{%
            \footnotesize 
            \hypertarget{link1}{\textsuperscript{1} \url{https://new.wikipedia.org/wiki/}} \\
            \hypertarget{link2}{\textsuperscript{2} \url{https://bloomlibrary.org/language:new}}%
        }%
}

\begin{table}[H]
    \centering
    \small
    \caption{Nwāchā Munā Corpus Overview}
    \label{tab:text_corpus_overview}
    \renewcommand{\arraystretch}{1.2}
    \begin{tabularx}{\columnwidth}{X r} 
        \hline
        \textbf{Metric} & \textbf{Count} \\
        \hline
        Total Sentences  & 5,727 \\
        Total Words & 27,644 \\
        Unique Words in Corpus & 8,599 \\
        \hline
    \end{tabularx}
\end{table}

\begin{table}[H]
    \centering
    \small
    \caption{Sentence Length Distribution in Nwāchā Munā Corpus}
    \label{tab:text_corpus_length}
    \renewcommand{\arraystretch}{1.5}
    \begin{tabularx}{\columnwidth}{X r} 
        \hline
        \textbf{Sentence Length} & \textbf{Count} \\
        \hline
        Short (0--5 words) & 4,323 \\
        Medium (6--10 words) & 1,073 \\
        Long ($>10$ words) & 331 \\
        \hline
    \end{tabularx}
\end{table}

\subsubsection{Audio Data Acquisition} 
Primary acoustic data was gathered through a dual-modal strategy involving original field recordings and the transcription of existing web-based audio, ensuring the model's exposure to varied acoustic environments. To maintain technical consistency across the dataset, all audio was standardized to a 16 KHz sampling rate and stored in a mono-channel WAV format, providing the necessary fidelity for deep learning-based feature extraction. For the recording phase, 18 native speakers from Banepa, Dhulikhel, Panauti, and Patan volunteered based on criteria such as native fluency and linguistic knowledge. While Newari exhibits location-based dialectal variations, the use of a standardized text across all recordings ensured that the resulting differences remained primarily tonal. To enhance the model's robustness against speaker-dependent variability, the data was stratified by age and gender. Audio recordings were captured using built-in smartphone microphones in an open environment with minimal acoustic disturbance. This field collection of 4 hours and 21 minutes of speech was further supplemented by approximately one hour of web-sourced audio \citep{meelen2024end}, which was originally in Romanized form and subsequently transliterated into Devanagari script by community members. 

The overall characteristics of the Nwāchā Munā speech corpus are summarized in Tables \ref{tab:corpus_overview}-\ref{tab:gender_distribution}. Table \ref{tab:corpus_overview} presents the corpus-level statistics, including the total number of utterances, total duration, and mean utterance length. Table \ref{tab:age_distribution} details the age distribution of the same participant set, while Table \ref{tab:gender_distribution} provides the gender distribution of the 18 speakers contributing to the corpus.

\begin{table}[H]
    \centering
    \small
    \caption{Overview of the Nwāchā Munā}
    \label{tab:corpus_overview}
    \renewcommand{\arraystretch}{1.6} 
    \begin{tabularx}{\columnwidth}{X r}
        \hline
        \textbf{Metric} & \textbf{Value} \\
        \hline
        Total Utterances & 5,727 \\
        Total Duration & 5.39 hours \\
        Mean Utterance Length & 3.39 sec \\
        \hline
    \end{tabularx}
\end{table}

\begin{table}[H]
\centering
\small
\caption{Age Distribution of Speakers in the Nwāchā Munā Corpus}
\label{tab:age_distribution}
\renewcommand{\arraystretch}{1.6} 
\begin{tabularx}{\columnwidth}{X r}
\hline
\textbf{Age Group} & \textbf{Number of Speakers} \\
\hline
16--25 & 6 \\
26--30 & 2 \\
30+ & 10 \\
\hline
Total Speakers & 18 \\
\hline
\end{tabularx}
\end{table}

\begin{table}[H]
\centering
\small
\caption{Gender Distribution of Speakers in the Nwāchā Munā Corpus}
\label{tab:gender_distribution}
\renewcommand{\arraystretch}{1.6} 
\begin{tabularx}{\columnwidth}{X r}
\hline
\textbf{Gender} & \textbf{Number of Speakers} \\
\hline
Male & 10 \\
Female & 8 \\
\hline
Total Speakers & 18\\
\hline
\end{tabularx}
\end{table}

\subsection{Model Training Strategies}

We adopted a comparative experimental framework to systematically evaluate cross-lingual transfer strategies for Newari automatic speech recognition. Our central hypothesis is that acoustic and orthographic proximity between Nepali and Newari enables effective encoder reuse, thereby reducing the need for large multilingual pretraining. To test this hypothesis, we structured our investigation into three core components: 1) acoustic modeling, 2) language modeling and decoding, and 3) semi-supervised learning.

In the domain of acoustic modeling, we evaluated distinct training strategies starting with a zero-shot evaluation of the pre-trained \texttt{NepConformer} \citep{poudel2025nepconformer}, followed by supervised fine-tuning of NepConformer, and fine-tuning of a massive multilingual model, \texttt{Whisper-Small} (244M parameters) on the collected Newari dataset. To test the adaptability of pre-trained acoustic representations, we further distinguished between standard full model fine-tuning and a decoder-only fine-tuning strategy. We also investigated the impact of data augmentation techniques in improving the performance and robustness of the models.

Finally, to address low-resource constraints in decoding, we explored language modeling strategies by incorporating an external KenLM-based n-gram model via shallow fusion to enhance transcription accuracy.

\subsubsection{Acoustic Modeling Strategies}

The zero-shot evaluation of NepConformer served as a baseline to assess cross-lingual generalization without any exposure to Newari during training. This setup enabled us to measure the extent to which acoustic and linguistic representations learned from Nepali transfer to a closely related but distinct language. Subsequently, we employed transfer learning through supervised fine-tuning on the Newari corpus to adapt the NepConformer model to the target language. This allowed us to measure the gains obtained from domain-specific adaptation compared to zero-shot performance. To test the adaptability of pre-trained acoustic representations, we further distinguished between standard full model fine-tuning and a decoder-only fine-tuning strategy, where the encoder parameters were frozen to assess the sufficiency of source-language acoustic features. Finally, we fine-tuned Whisper-Small to evaluate whether broad cross-lingual pre-training yielded stronger transfer capabilities than a monolingual Nepali trained model.

To combat data scarcity in low-resource Nepal Bhasha ASR, a dual-stage data augmentation strategy was employed. By combining static (offline) and dynamic (online) methods, we introduced diverse acoustic variations to improve model robustness and prevent overfitting. This approach acts as a regularizer, forcing the model to learn invariant phonetic features rather than memorizing the specific acoustic signatures of the limited training set.

\subsubsection{Language Modeling and Decoding}
To enhance linguistic modeling under low-resource constraints, we used two approaches: decoder-only fine-tuning and shallow fusion with an external language model (KenLM) \cite{heafield-2011-kenlm}.

In the decoder-only setup, the encoder parameters were frozen, and only the prediction network and joint network were updated during fine-tuning. By applying the phonetic similarities between Nepali and Newari, this strategy conserves the transferable acoustic representations obtained during pre-training while allowing the decoder to adapt specifically to Newari linguistic patterns. 

Independent of decoder adaptation, we also integrated an external n-gram language model trained using KenLM and applied shallow fusion during inference. This introduces explicit linguistic priors to capture word-level dependencies that the limited paired data might miss. During beam search decoding, the log-probabilities from the acoustic model and the external language model were linearly combined~\citep{toshniwal2018comparison}, with the fusion weight tuned on the development set.

For the baseline system, beam search decoding approximates the most likely sequence:
\begin{equation*}
y^* = \arg\max_{y} \log p(y \mid x)
\end{equation*}

Under shallow fusion, we incorporate the external language model via log-linear interpolation with a word insertion bonus:
\begin{equation*}
y^* = \arg\max_{y} \left\{ \log p(y \mid x) + \alpha \log p_{LM}(y) + \beta |y|_w \right\}
\end{equation*}
where $\alpha$ controls the language model weight, $\beta$ is the word insertion bonus, and $|y|_w$ denotes the number of words in the hypothesis $y$.

Decoder-only fine-tuning improves the model’s language understanding by updating only the higher-level components while keeping the encoder fixed, which also reduces computational cost. This setup allows us to observe the effect of adapting the decoder alone in a low-resource setting. While, shallow fusion adds an external language model during decoding without changing the acoustic model itself. By evaluating these two methods independently, we can clearly measure the contributions of internal decoder adaptation and external statistical language modeling to CER, WER, and improve transcription consistency in low-resource settings.

\subsubsection{Semi-Supervised Learning}

Due to the availability of limited data for Nepal Bhasha, we adopted a semi-supervised pseudo-labeling framework. An unlabeled corpus was collected from publicly available broadcast sources, including radio, podcasts, and television, and segmented into shorter utterances. To mitigate memory issues and alignment degradation in CTC-based training, excessively long utterances were discarded. The remaining audio was transcribed using our best intermediate acoustic model to generate pseudo-labels.

To ensure quality, we applied rigorous filtering. Single-word transcripts were removed, as they were typically noise-induced fragments rather than valid words. We also computed average prediction confidence scores and applied a threshold filter to reduce error reinforcement before incorporating the pseudo-labeled data into our training pipeline.


\section{Results and Discussion}
\subsection{Experimental Setup}

The dataset was divided into 80\% for training, 10\% for validation, and 10\% for testing. Due to the limited size of the dataset across 18 participants, speaker representation was maintained in all splits to prioritize overall acoustic diversity rather than enforcing strict speaker independence. Table \ref{tab:dataset_split} summarizes the Newari speech corpus split across training, validation, and test sets, showing the number of hours and utterances for each partition.

\begin{table}[H]
\centering
\caption{Dataset split for Newari speech corpus}
\label{tab:dataset_split}
\renewcommand{\arraystretch}{1.25}
\begin{tabular}{lcccc}
\hline
\textbf{Split} & \textbf{Hours} & \textbf{Utterances} \\
\hline
Train & 4.31 & 4575 \\
Validation & 0.54 & 576 \\
Test & 0.54 & 576 \\
\hline
\end{tabular}
\end{table}

For the Conformer model, static augmentation expanded the training set by a factor of 5 via speed perturbation ($0.9\times, 1.1\times$), volume randomization, and noise injection, resulting in 23.05 total hours of audio. Dynamic augmentation was applied with a probability of $p=0.5$, involving time stretching (rate $0.8\times$--$1.25\times$), pitch shifting  ($\pm 4$ semitones), and Gaussian noise injection (amplitude $0.001$--$0.015$).

The fine-tuned Nepal Bhasha ASR model utilizes a  \texttt{Conformer-CTC }architecture with a Byte Pair Encoding (BPE) tokenizer of vocabulary size 128. The 18-layer Conformer encoder has a d-model dimension of 256, 4 attention heads, and a convolution kernel size of 31. Regularization includes a dropout rate of 0.1 across the encoder and attention modules, alongside SpecAugment with 2 frequency masks (width 27) and 2 time masks (width 70). Optimization is performed using AdamW with a learning rate of $1 \times 10^{-4}$ and a Cosine Annealing schedule with 3,000 warmup steps. T4 GPU with 16GB vRAM and 29GB of system RAM was used for training in Kaggle environment with total training time of 3 hours for standard data and 12 hours for augmented data.

The \texttt{Whisper-Small} model was fine-tuned for 10 epochs using a sequence-to-sequence objective, with the decoder explicitly forced to the Nepali language token in order to leverage script compatibility. Optimization was performed using AdamW with a cosine learning rate decay (peak $1 \times 10^{-5}$, 10\% warmup) and an effective batch size of 16, achieved via gradient accumulation under FP16 mixed-precision. To further ensure stability, a dropout rate of 0.2 to all attention and hidden layers were applied. A T4-GPU with 15GB vRAM was used for training this model in Google Colab environment with total training time of 4.5 hours.

For language modeling and decoding, we performed a grid search on the development set to identify optimal decoding hyperparameters across all models. We explored different configurations of language model weight ($\alpha$), word insertion bonus ($\beta$), and beam width. Based on this systematic evaluation, a 5-gram KenLM with $\alpha=0.4$, $\beta=1.5$, and a beam width of 128 was selected, as it consistently yielded the best performance improvements. The language model was trained on a text corpus consisting of approximately 51k utterances and 478k word tokens, sourced from Nepal Bhasha Wikipedia. Notably, this dataset was reserved solely for language modeling and is distinct from the audio transcription data.

For the semi-supervised learning experiments, an initial unlabeled corpus of 13.65 hours was collected. During the filtering phase, utterances longer than 15 seconds were discarded. After transcribing the audio with our intermediate acoustic model, a 70\% confidence threshold was applied to the predictions to ensure pseudo-label quality. Following this rigorous filtering process, 9.33 hours of high-quality pseudo-labeled data remained for experimental training.

\subsection{ Findings and Analysis }
The performance of the models was evaluated using CER as the primary metric. The experiments were conducted on the Nwāchā Munā dataset, while additional evaluations incorporated augmented data and unlabelled speech to examine the effect of data expansion and semi-supervised training. The quantitative results are summarized in Table \ref{tab:results}, where only the best-performing models at their optimal training epochs are reported. 

\begin{table*}[!b]
\centering
\small
\caption{Summary of Results for Evaluated Modeling Strategies}
\label{tab:results}
\renewcommand{\arraystretch}{1.25}
\begin{tabularx}{\textwidth}{>{\hsize=1.3\hsize}X >{\hsize=0.7\hsize}X r r}
\hline
\textbf{Strategies (model details)} & \textbf{Dataset} & \textbf{CER (\%)} & \textbf{WER (\%)} \\
\hline
Zero-Shot NepConformer & - & 52.54 & 98.68 \\
Semi-supervised NepConformer  & 5.39 hours labelled \newline + 9.33 hours Pseudo labeled & 19.83 & 58.28\\
NepConformer + 5-gram KenLM (Shallow fusion, $\lambda=0.4$, $\beta=1.5$, beam 128) & 5.39 hours labelled & 19.75 & \textbf{48.46} \\
Decoder-only NepConformer (Encoder frozen)  & 5.39 hours labelled & 18.77 & 53.47 \\
Whisper-Small (Fine-tuned on base data) & 5.39 hours labelled & 18.76 & 51.44 \\
NepConformer (Fine-tuned on base data) & 5.39 hours labelled & 18.72 & 54.34 \\
Whisper-Small + Augmentation (Time stretch, pitch shift, Gaussian noise) & 5.39 hours labelled & 17.88 & 48.80\\
NepConformer + Augmented Data (speed perturbation ($0.9\times, 1.1\times$), volume randomization, and noise injection) & 23.05 hours labelled & \textbf{17.59} & 52.61\\
\hline
\end{tabularx}
\end{table*}

Due to differences in script representation, normalization, evaluation protocol and the unavailability of their exact test split, a direct numerical comparison with \cite{meelen2024end} is not feasible. Instead, our experimental results demonstrate that transferring representations from a neighboring language with a similar script is a viable alternative to massive multilingual pre-training for low-resource ASR, suggesting that linguistic proximity can potentially outweigh model scale in ultra-low-resource South Asian scenarios. 

Table \ref{tab:ortho_comparison_single} shows the orthographic and phonological comparison of the Nepali and Newari Language. Table \ref{tab:computation_parameter} presents a comparison of model size and computational resource requirements, highlighting that \texttt{NepConformer} is significantly more lightweight, with only 30.54 million parameters and lower VRAM usage (8.56 GB) compared to \texttt{Whisper-Small}, which requires substantially higher usage (12.38 GB). 

Despite possessing significantly fewer parameters and lacking the huge data backing available in multilingual models like Whisper, the \texttt{NepConformer} model achieved a baseline CER of 18.72\%, effectively matching the performance of the fine-tuned \texttt{Whisper-Small} model (18.76\%). This parity suggests that the acoustic and orthographic overlap between Nepali and Newari facilitates robust feature extraction, minimizing the need for the extensive capacity of large-scale models. Furthermore, the application of data augmentation proved critical in this data-scarce regime, yielding the state-of-the-art performance of 17.59\% CER for \texttt{NepConformer} and 17.88\% for \texttt{Whisper-Small}, highlighting that augmenting scarce audio data can effectively overcome dataset limitations and drive substantial performance gains.

\begin{table}[h]
\centering
\small
\caption{Orthographic and Phonological Comparison of Nepali and Newari}
\label{tab:ortho_comparison_single}
\renewcommand{\arraystretch}{1.5}
\begin{tabularx}{0.9\columnwidth}{X X X}
\hline
\textbf{Features} & \textbf{Nepali} & \textbf{Newari} \\
\hline
Script & Devanagari & Devanagari \\
Nasalization Markers & Yes & Yes (more enhanced) \\
Agglutination & Moderate & High \\
Halant Usage & Standard & High Morphological Load \\
\hline
\end{tabularx}
\end{table}

\begin{table}[h]
\centering
\small
\caption{Computational Resource and Parameter Comparison}
\label{tab:computation_parameter}
\renewcommand{\arraystretch}{1.5}
\begin{tabularx}{0.9\columnwidth}{X X X}
\hline
\textbf{Model} & \textbf{Parameter (Million)} & \textbf{VRAM Used (GB)} \\
\hline
NepConformer & 30.54 & 8.56 \\
Whisper(small) & 244 & 12.38 \\
\hline
\end{tabularx}
\end{table}

Conversely, a closer examination of our training configurations reveals critical insights into the adaptation process. The poor zero-shot performance (52.54\% CER) confirms that despite script sharing, the phonological distinctiveness of Newari requires explicit fine-tuning. The \textit{decoder-only} fine-tuning of \texttt{NepConformer} yielded results (18.77\% CER) nearly identical to full model fine-tuning, indicating that the pre-trained Nepali encoder features are sufficiently generalized to encode Newari speech without modification. 

Additionally, the integration of the KenLM n-gram language model was evaluated across all models during decoding. The best results were observed with the NepConformer fine-tuned model, where WER was reduced by approximately 11.7\% relative to the baseline. At the same time, the CER increased slightly by 1.37\%, as the language model tends to favor standard spellings, which can override phonetically correct but unusual or local forms in Newari speech. This happens partly because the text used to train the language model doesn’t always match how people actually speak. Overall, this shows the trade-off between making words more consistent and keeping the natural variations of this low-resource, agglutinative language.

Finally, in the semi-supervised learning approach, incorporating the pseudo-labeled data with the original labeled training set degraded performance, increasing the CER to 19.83\% compared to the 17.59\% baseline. A qualitative analysis of a random sample of these pseudo-labels revealed three primary sources of error: extensive code-mixing (alternating with Nepali or English), frequent overlapping conversational speech, and highly variable acoustic environments (e.g., background music, distant microphones). Ultimately, this decline highlights the critical challenge that the severe domain shift introduced by uncurated data can completely overshadow the benefits of increased training volume. This suggests that strict domain alignment, speaker diarization, and rigorous filtering are significantly more crucial than raw data quantity for endangered-language ASR.

\subsection{Error Analysis}

Newari’s rich morphological structure, characterized by productive affixation and concatenated morphemes, makes character-level analysis particularly informative. Although the model frequently predicts individual characters correctly, it struggles to assemble them into accurate representations. As a result, the CER remains relatively low, while the WER is substantially higher, reflecting the difficulty of modeling agglutinative formations and complex orthographic concatenations. This is also observed in other work~\citep{k-etal-2025-advocating}.

Qualitative analysis indicates that transcription errors are primarily associated with specific diacritics: {\nep हलन्त} ({\nep \char"25CC}{\nep ्}) (vowel suppressor), {\nep अनुस्वार} ({\nep \char"25CC}{\nep ं}) (nasalization), {\nep चंद्रबिंदु} ({\nep \char"25CC}{\nep ँ}) (nasal vowel), and {\nep विसर्ग} ({\nep \char"25CC}{\nep ः}) (aspiration). These markers encode essential phonetic distinctions, including nasalization, short stops, and breathy releases; their omission alters the acoustic-orthographic correspondence. As further illustrated in Table \ref{tab:error_results}, these findings underscore the importance of language-specific grapheme modeling and the development of corpora encompassing both formal and conversational usage.\\

\begin{table}[H]
\centering
\small
\caption{Error Frequency Table}
\label{tab:error_results}
\renewcommand{\arraystretch}{1.5} 
\begin{tabularx}{0.9\columnwidth}{X r}
\hline
\textbf{Error Type} & \textbf{\% of total errors} \\
\hline
Word Boundary Errors & 35\% \\
Halant/Cluster Errors & 25\% \\
Nasalization \& Anusvara Confusion & 20\% \\
Lexical Substitution & 15\% \\
Function Word \& Particle Error & 5\%\\
\hline
\end{tabularx}
\end{table}

\begin{table*}[t]
\centering
\caption{Qualitative Error Analysis of Nwāchā Munā (underline in prediction column represents the error segment)}
\label{tab:error_analysis}
\footnotesize
\renewcommand{\arraystretch}{1.5}
\begin{tabularx}{\textwidth}{l X X X}
\hline
\textbf{No.} & \textbf{Reference} & \textbf{Prediction} & \textbf{Error Analysis} \\
\hline

(i) & 
{\nep तस्सकं न्ह्याइपुसेच्वं सीम बत्तिं चैं छन्दत म्याहला च्वंगु} \newline
tassakaṃ nhyāipusecvaṃ sīma battiṃ caiṃ chandata myāhalā cvaṃgu \newline
\textit{It was a lot of fun sitting even in the dim light, singing your song}
& 
{\nep तस्स\underline{कां} \underline{न्येयेपुसं} \underline{सिमा}\underline{पत्ति} चैं \underline{छंत} \underline{मां ला} च्वं} \newline
tassa\underline{kāṃ} \underline{nyeyepusaṃ} \underline{simāpatta} caiṃ \underline{chaṃta} \underline{māṃ lā} cvaṃ
& 
Phonetic substitutions and word distortions; nasal markers and vowels altered within agglutinated structures. \\ \hline

(ii) & 
{\nep दइ धका नं सिल हे} \newline 
daï dhakā naṃ sila he
& 
{\nep दइ ध\underline{काः} नं सिल हे} \newline 
daï dha\underline{kāḥ} naṃ sila he
& 
Insertion of {\nep (ः)} (Visarga), reflecting minor phonetic misalignment. \\ [-3ex]
 & \textit{They say it is known that it will fall} \\

\hline

(iii) &
{\nep फोहोर जुलकि लाकां सिले माः} \newline
phohora julaki lākāṃ sile māḥ \newline
\textit{If the shoe is dirty, it must be washed}
&
{\nep \underline{पर} जुलकि \underline{लाका} सि\underline{ने} माः} \newline
\underline{para} julaki \underline{lākā} si\underline{ne} māḥ
&
Nasal suffix deletion and lexical substitution; final {\nep (ं)} dropped, with minor token distortion.\\
\hline

(iv) & 
{\nep आः थ्व सु} & 
{\nep \underline{आाख्वसु}} &
Word boundary deletion; independent tokens merged due to agglutinative structure. \\[-5ex]
& āḥ thva su &
\underline{āākhvasu} & \\ [-0.5ex]
& \textit{Now, who is this}
\\
\hline
\end{tabularx}
\end{table*}

The sample qualitative error are presented in Table \ref{tab:error_analysis}. This distribution shows that the primary challenges arise not from isolated character mis-recognition but from morphological segmentation complexity, diacritic modeling, and the compounding effect of Devanagari’s script characteristics.


\section{Conclusion}
In this work, we developed resources to help bridge the digital divide for Nepal Bhasha (Newari) by curating a high-quality 5.39-hour transcribed speech corpus, with all transcripts encoded in the Devanagari script. Using this resource, we systematically evaluated cross-lingual transfer learning strategies, benchmarking \texttt{NepConformer} along with a multilingual \texttt{Whisper-Small} baseline. Our experiments demonstrate substantial gains from supervised fine-tuning and data augmentation, reducing the CER from a high zero-shot baseline to below 18\%. The frozen-encoder fine-tuning strategy further enabled stable and effective adaptation in this ultra-low-resource setting. Moreover, shallow fusion with a KenLM $n$-gram language model enhances lexical regularity and contributes to a reduction in WER, even when character-level improvements remain marginal, indicating improved word-level consistency in decoded outputs.
Beyond Nepal Bhasha, our findings indicate that intra-regional transfer learning within South Asian language clusters may offer a computationally efficient pathway for scaling ASR to other endangered and minority languages

\paragraph{Data and Code Availability}
The training scripts and source code are available on GitHub  (\url{https://github.com/ilprl/nwacha-muna}). The model weights and audio datasets are hosted on Hugging Face (\url{https://huggingface.co/collections/ilprl-docse/nwacha-muna}).

\section{Ethics Statement and Limitations}
This work offers meaningful societal and technological benefits for the Nepal Bhasha speaking community. By providing an open-source Devanagari speech corpus and establishing strong ASR benchmarks, the resources developed pave the way for voice-driven AI products that can significantly enhance digital accessibility. Furthermore, this research plays an important role in building extensive speech archives, allowing endangered language communities to actively document and preserve their linguistic heritage. 

The dataset was collected through voluntary participation from the community, involving native speakers across diverse age groups and genders to ensure representative coverage. For the recorded audio,  consent was obtained from all volunteers, and no sensitive or personally identifiable information (PII) was intentionally included in the spoken transcripts. Nonetheless, we acknowledge that models trained on limited data may reflect embedded societal biases or result in acoustic-phonetic misrepresentations, particularly in ultra-low-resource settings. Because of these limitations, the baseline ASR system developed in this study should not be deployed in critical applications without rigorous human verification. The primary objective of this work remains the promotion of linguistic inclusion and digital support for under-resourced language communities.

Despite the promising benefits, this work faces several constraints related to ultra-low-resource settings. Primarily, the curated Nwāchā Munā dataset, while high-quality, remains relatively small compared to standard ASR benchmarks. Our current evaluation focuses predominantly on read speech and formal sentences, which may not fully capture the acoustic variability, disfluencies, and rapid speaking rates encountered in everyday spontaneous conversations. Also, external n-gram language model utilized during shallow fusion is constrained by domain mismatches and possesses limited coverage of natural, conversational linguistic patterns.

\section{Bibliographical References}
\label{sec:reference}

\bibliographystyle{lrec2026-natbib}
\bibliography{NewaConformer-references}

\end{document}